\documentclass[letterpaper, 10 pt, conference]{ieeeconf}  

\IEEEoverridecommandlockouts                              

\usepackage{graphicx}
\usepackage{amsmath}
\usepackage{algorithm}
\usepackage{algpseudocode}
\usepackage{xcolor}
\usepackage{soul}
\usepackage{stfloats}
\usepackage{amsfonts} 
\usepackage{amssymb}
\usepackage{comment}
\usepackage[tight]{subfigure}
\renewcommand{\baselinestretch}{1}
\usepackage{xcolor}
\usepackage{comment}
\usepackage{hyperref}
\newcommand\edit[1]{{\color{black}#1}}
\usepackage{indentfirst}

\title{\LARGE \bf
Faithful Euclidean Distance Field from Log-Gaussian Process Implicit Surfaces
}
\author{Lan Wu, Ki Myung Brian Lee, 
Liyang Liu, Teresa Vidal-Calleja \thanks{This work was supported by the Australian Government Research Training Program and the Australian Government Department of Agriculture \& Water Resources as part of its Rural R\&D for Profit programme, MLA under Grant V.RDP.2005.}\thanks{All authors are with the Centre for Autonomous Systems at the Faculty of Engineering and IT, University of Technology Sydney, Australia.
Corresponding author: \texttt{Lan.Wu-1@student.uts.edu.au.}%
}
}

\begin{document}

\maketitle
\thispagestyle{empty}
\pagestyle{empty}

\begin{abstract}
In this letter, we introduce the Log-Gaussian Process Implicit Surface (Log-GPIS), a novel continuous and probabilistic mapping representation suitable for surface reconstruction and local navigation. \edit{Our key contribution is the realisation that the regularised Eikonal equation can be simply solved by applying the logarithmic transformation to a GPIS formulation to recover the accurate Euclidean distance field~(EDF) and, at the same time, the implicit surface.} To derive the proposed representation, \emph{Varadhan's formula} is exploited to approximate the non-linear Eikonal partial differential equation (PDE) of the EDF by the logarithm of a linear PDE. We show that members of the Mat\'{e}rn covariance family directly satisfy this linear PDE. The proposed approach does not require post-processing steps to recover the EDF. Moreover, unlike sampling-based methods, Log-GPIS does not use sample points inside and outside the surface as the derivative of the covariance allow direct estimation of the surface normals and distance gradients. We benchmarked the proposed method on simulated and real data against state-of-the-art mapping frameworks that also aim at recovering both the surface and a distance field. Our experiments show that Log-GPIS produces the most accurate results for the EDF and comparable results for surface reconstruction and its computation time still allows online operations.


\end{abstract}

\section{Introduction}

Autonomous robots navigating in unknown and unstructured environments require accurate representations of the world. These representations can be used for different purposes such as localisation, mapping,  path planning, manipulation, amongst others. Usually, each application requires its dedicated map representation. For instance, navigation requires the robot to build an efficient representation with distance and direction to collision  information  that  are  suitable  for  local  or  global path  planning.  On the other hand, recovering  a  high-quality surface reconstruction of the scene for mapping purposes, requires dense, accurate and high-resolution representation. In this work, we are interested in world representations that can be used concurrently for accurate reconstruction of the scene and for local navigation. 

A wide variety of mapping representations have been proposed in the literature that are suitable for navigation~\cite{OccupancyMapsMoravecElfes, Octomap, o2012gaussian, Chomp} or for surface reconstruction~\cite{kinectfusion, LanRAL20, paper:GeomPrior, GPmap}. Notably, occupancy maps for navigation and path planning have received plenty of attention over the years. Occupancy maps vary from discrete and probabilistic 2D~\cite{OccupancyMapsMoravecElfes} and 3D~\cite{Octomap} to continuous and also probabilistic~\cite{o2012gaussian,GPmap} representations. On the other hand, a widely adopted mapping representation for surface reconstruction in robotics is the truncated signed distance field~(TSDF) used in Kinect fusion~\cite{kinectfusion} to recover a 3D mesh of the scene. Although some authors have exploited it for navigation~\cite{TSDFplanning1}~\cite{TSDFplanning2}, TSDF is more suitable as an implicit representation of the surface than as a navigation distance field, given that it is an approximation of the Euclidean signed distance field~(ESDF)~\cite{Helen2016signed}. 

Distance fields are a natural representation for navigation and planning and have been proven advantageous in the past~\cite{Chomp}. In fact, Oleynikova \emph{et al.} in~\cite{Helen2016signed} show that distance fields can be used for both surface reconstruction and planning seamlessly. In~\cite{Voxblox} the same authors proposed a discrete mapping framework called Voxblox based on distance fields, which recovers the surface reconstruction of a scene with TSDFs and uses an approximation of the ESDF for local path planning. Despite the fact that Voxblox requires conversion from TSDF to ESDF, the framework runs efficiently online. 

\begin{figure}[t]
  \centering
  \resizebox{\linewidth}{!}{
  \subfigure[Ground truth]{\includegraphics[height=3.3cm]{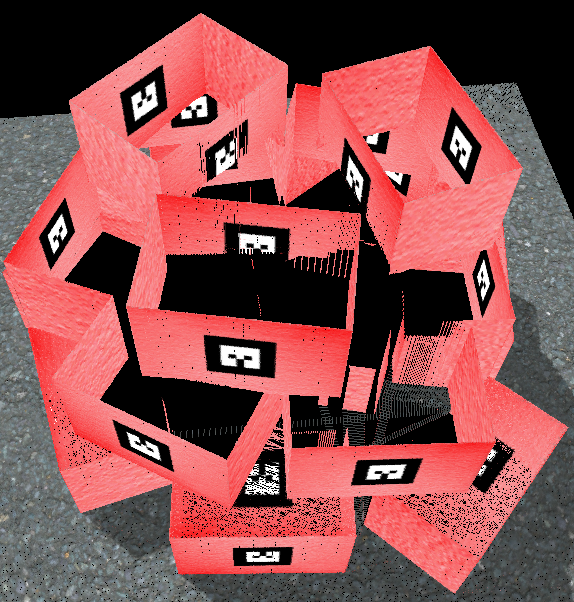}}
  \subfigure[Our result]{\includegraphics[height=3.3cm]{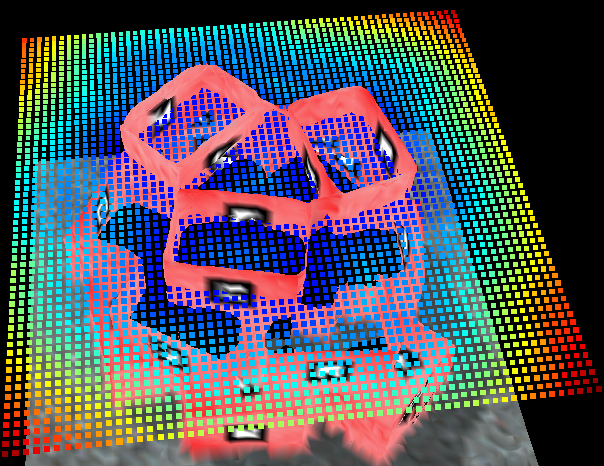}}
  }
  \caption{3D reconstruction of a pile of boxes in a simulated environment with (a) the ground truth and (b) the surface reconstruction using Log-GPIS (mesh) and a slice of the Euclidean distance field (point cloud).}
  \label{teaser}
\end{figure}

Inspired by the idea of distance fields as a unified representation for surface reconstruction and navigation, our work here proposes a single representation that has the property of directly inferring both, the implicit surface and the accurate Euclidean distance field of the scene (see Fig.~\ref{teaser}b)). Moreover, unlike most TSDF-based approaches, our proposed representation is continuous and probabilistic. 


Gaussian process implicit surface (GPIS)~\cite{Microsoft},~\cite{Bhoram} is a recently proposed probabilistic and continuous representation of implicit surfaces that exploits distance information. However, with GPIS, the distance information is not accurate far away from the surface due to the lack of training points. Our work addresses this issue by faithfully modelling a fundamental property of Euclidean distance fields (EDFs) inspired by the proposed heat method~\cite{Crane}. 

We anticipate that Log-GPIS will be pivotal in the adoption of GPIS in practical robotic mapping and planning applications. As we show, its mathematical foundation provides significant improvement in accuracy and fidelity, while requiring only minor modifications to standard GPIS\edit{~\cite{Bhoram}}. We believe that the combination of smooth, probabilistic nature of GPIS and the improved accuracy afforded by our method will unlock the potential of GPIS as a unified representation for robotic mapping and planning. 
The contributions of this paper are as follow:

\begin{itemize}
\item A novel continuous and probabilistic representation that directly infers the EDF and the implicit surface with gradients. Given any query point in space, our representation provides the Euclidean distance and direction to the nearest surface.
\item The derivation of the Log-Gaussian Process Implicit Surface (Log-GPIS) by exploiting Varadhan's approach~\cite{Varadhan} to approximately enforce the Eikonal equation via a \edit{linear} partial differential equation (PDE). We show that by simply applying the logarithmic function to a GPIS with a specific covariance of the Mat\'{e}rn family, the regularised Eikonal equation is satisfied to accurately infer the EDF and concurrently the implicit surface.
\item An implementation suitable for online inference following~\cite{Bhoram} that can be used for local path planning, which employs the marching cubes algorithm in~\cite{fuhrmann2015accurate} to recover the surface mesh for visualisation.  


\end{itemize}
 
The remainder of the paper presents a brief background on EDF and GPIS in Section~\ref{sec:background} followed by the derivation of the Log-GPIS in Section~\ref{sec:approach}. The evaluation using simulated and real data is presented in Section~\ref{sec:evaluation} and the conclusions are in Section~\ref{sec:conclusion}.

\section{Background}\label{sec:background}
\subsection{Euclidean Distance Field}
Consider a manifold $S \subset \mathbb{R}^{D}$ with boundary $\partial S$, of which the orientation is given by the normal $\mathbf{n}$. 
For all $\mathbf{x} \in \mathbb{R}^{D}$, the EDF $d(\mathbf{x})$ is \edit{the closest distance to the boundary $\partial S$}:
\begin{equation}
    d(\mathbf{x})=\min_{\mathbf{y} \in \partial S} \edit{ | \mathbf{x} - \mathbf{y} | }
\end{equation}
Consider the scenario where the boundary $\partial S$ dilates along its normal direction with unit speed. 
Then, the arrival time will be equal to the distance $d(\mathbf{x})$. 
This intuition is captured by the Eikonal equation~\cite{Mauch}, which is given by
\begin{equation}\label{eikonal}
    |\nabla d\edit{(\mathbf{x})}|=1 \quad \mathbf{x} \in \mathbb{R}^{D}
\end{equation}
with boundary constraints,
\begin{equation}
d\edit{(\mathbf{x})}=0 \quad \text { and } \quad \partial d\edit{(\mathbf{x})} / \partial \boldsymbol{n}=1 \quad \text { on } \edit{\mathbf{x} \in} \partial S.\\
\end{equation}

Suppose we are given sparse measurements of the points on surface, $\mathcal{X} = \{ \mathbf{x}_{i} \} \subset \partial S$, $i = 1 \ldots N$. 
The aim of this paper is to estimate $d(\mathbf{x})$ given $\mathcal{X}$, thereby reconstructing $S$. 

\subsection{Gaussian Process Implicit Surfaces}\label{sec:background:gpis}
Gaussian process (GP) regression~\cite{GPbook} is a well-established non-parametric approach to solving non-linear regression problems. It allows estimating the value of an unknown function at an arbitrary query point given noisy and sparse measurements at other points. In this work, we consider zero-mean GPs:
\begin{equation}
    f \sim \mathcal{GP}(0, k(\mathbf{x}, \mathbf{x}')), 
\end{equation}
where $k(\mathbf{x}, \mathbf{x}')$ is the covariance function that characterises the properties of $f$. 

Suppose we are given a training dataset of measurements $\mathbf{y}=\left\{y_{i} = f(\mathbf{x}_{i}) + \epsilon_{i} \right\}_{i=1}^{N} \subset \mathbb{R}$, taken at locations $\mathbf{x}_{i} \in \mathbb{R}^{D}$ and corrupted by additive Gaussian noise $\epsilon_{i} \sim \mathcal{N}(0, \sigma_{f}^2)$. The posterior distribution of the value of $f$ at an arbitrary testing point $\mathbf{x}_{*}$ is given by $f(\mathbf{x}_{*}) \sim \mathcal{N}(\bar{f}_{*},\mathbb{V}\left[f_{*}\right])$, where the predictive mean and variance are given by:
\begin{equation}
\begin{aligned}
\bar{f}_{*}=\mathbf{k}_{*}^{\top}\left(K+\sigma_{f}^{2} I\right)^{-1} \mathbf{y}, 
\end{aligned}
\end{equation}
\begin{equation}
\mathbb{V}\left[f_{*}\right]=k\left(\mathbf{x}_{*}, \mathbf{x}_{*}\right)-\mathbf{k}_{*}^{\top}\left(K+\sigma_{f}^{2} I\right)^{-1} \mathbf{k}_{*}.
\end{equation}
Here, $I$ is the identity matrix, and $\mathbf{y}$ represents the input values. $\mathbf{k}_{*}$ is the vector of covariances between the $n_{x}$ training points and the testing point. $K$ is the $n_{x} \times n_{x}$ covariance matrix of the training points. $k\left(\mathbf{x}_{*}, \mathbf{x}_{*}\right)$ is the covariance function value of the testing point.

Furthermore, as mentioned above, GP is capable of predicting not only the function values, but also the gradients. 
This is because differentiation is a linear operator on the space of functions, and consequently the derivative of a GP is another GP~\cite{gp_derivative}. 
For $f \sim \mathcal{GP}(0, k(\mathbf{x}, \mathbf{x}'))$, the gradient GP $\nabla f$ is given by applying the gradient operator $\nabla$ on the covariance function,
\begin{equation}
    \begin{bmatrix}
        f \\
        \nabla f
    \end{bmatrix} \sim \mathcal{GP}(\mathbf{0}, \tilde{\edit{k}}(\mathbf{x}, \mathbf{x}')),
\end{equation}
where \edit{$\tilde{\edit{k}}(\mathbf{x}, \mathbf{x}')$ is the joint covariance matrix of $k(\mathbf{x}, \mathbf{x}')$ and the partial derivatives of $k(\mathbf{x}, \mathbf{x}')$ at $\mathbf{x}$ and $\mathbf{x}'$~\cite{paper:GeomPrior},}
\begin{equation}
    \tilde{\edit{k}}\left(\mathbf{x}, \mathbf{x}'\right)=\left[\begin{array}{cc}
k\left(\mathbf{x}, \mathbf{x}'\right) & k\left(\mathbf{x}, \mathbf{x}'\right) \nabla_{\mathbf{x}'}^{\top}  \\
\nabla_{\mathbf{x}} k\left(\mathbf{x}, \mathbf{x}'\right) & \nabla_{\mathbf{x}} k\left(\mathbf{x}, \mathbf{x}'\right) \nabla_{\mathbf{x}'}^{\top}
\end{array}\right].
\end{equation}

\edit{Correspondingly, given the training dataset including the observations \edit{$\mathbf{y}$} and its gradients $\nabla \mathbf{y}$, where the gradients are computed in the same way as in~\cite{Bhoram}}, the following equations express the mean and variance at any testing point with \edit{gradient} inference,
\begin{equation}\label{eq:gp_inference_gradients}
\begin{aligned}
\begin{bmatrix}
\bar{f}_{*} \\
\nabla \bar{f}_{*}
\end{bmatrix} &=
 \tilde{\mathbf{k}}_{*}^{\top}(\tilde{K}+\Sigma_{f}^2 I)^{-1}\begin{bmatrix}
\mathbf{y} \\
\nabla \mathbf{y}
\end{bmatrix}
\end{aligned}
\end{equation}
\begin{equation}
\begin{aligned}
\begin{bmatrix}
\mathbb{V}\left[f_{*}\right] \\
\mathbb{V}\left[\nabla{f}_{*}\right]
\end{bmatrix} &=
 \tilde{k}(\mathbf{x}_{*}, \mathbf{x}_{*})-
 \tilde{\mathbf{k}}_{*}^{\top}(\tilde{K}+\Sigma_{f}^2 I)^{-1}\tilde{\mathbf{k}}_{*}.
\end{aligned}
\end{equation}

GP implicit surface (GPIS) techniques~\cite{Microsoft},~\cite{Bhoram},~\cite{stork2020ensemble}~\cite{LanRAL20} use GP regression to estimate the distance field of the surface. Consequently, the surface is given by the zero-level set of the distance field. Most \edit{standard} GPIS techniques directly model the distance field as a GP~\edit{\cite{Microsoft}}, and often use normal measurements to ensure correct results\edit{~\cite{Bhoram}}. However, while this allows continuous and probabilistic representation of the surface with high precision and continuity \edit{of the distance field} near the surface, 
it is often highly inaccurate further from the surface. For example, using a zero-mean GP implies that the prediction will include false-positive artifacts as the predicted function value falls back to zero. Our method, \emph{Log-GPIS}, alleviates this issue, by elegantly enforcing the Eikonal equation~\eqref{eikonal} into GP regression. 



\section{\edit{Log-Gaussian Process Implicit Surface}}\label{sec:approach}
\edit{
\subsection{Method}
As the name suggests, Log-GPIS simply requires that one takes the logarithm of the standard GP regression in~\eqref{eq:gp_inference_gradients} to produce accurate EDF estimates near and far away from a surface sampled with a 2D or 3D depth sensor. 

We consider GPIS as described in Sec.~\ref{sec:background:gpis}, equipped with the Mat\'{e}rn family of covariance functions:
\begin{equation}\label{eq:mattern}
    k(\mathbf{x}, \mathbf{x}^{\prime})=\sigma^{2} \frac{2^{1-\nu}}{\Gamma(\nu)}\left(\frac{\sqrt{2 \nu}}{\lambda} |\mathbf{x}-\mathbf{x}^{\prime}|\right)^{\nu} K_{\nu}\left(\frac{\sqrt{2 \nu}}{\lambda} |\mathbf{x}-\mathbf{x}^{\prime}|\right),
\end{equation}
where the characteristic length scale $\lambda$  and variance $\sigma^{2}$ are hyperparameters of the GP model. $K_{\nu}$ is the modified Bessel function of the second kind of order $\nu$.

Given a set of points on the surface, $\mathcal{X} = \{\mathbf{x}_{i}\}$, the distance $\bar{d}_{*}$ at a query point $\mathbf{x}^{*}$ is obtained by taking the logarithm of the GP inference result with target measurement set as $y_{i} = 1$, 
\begin{equation}
    \bar{d}_{*}=-\frac{\ln \bar{f}_{*}}{\lambda},
    \label{3}
\end{equation}
where $\bar{f}_{*}$ is the predictive mean from GP inference~\eqref{eq:gp_inference_gradients}.
An immediate benefit of the log-transformation is that the predicted distance approaches infinity as we query points further away from the surface, while the standard GPIS~\cite{Bhoram} will predict a distance value around zero, resulting in undesirable artifacts. 
Further, note that for Log-GPIS distance value of $\log(1) = 0$ only occurs on the surface boundary.
Thus, Log-GPIS solves the issue of undesirable artifacts in the predicted surface geometry.

The log-transformation also affects the gradient. Taking the gradient of both sides of \eqref{3} with respect to the distance, 
\begin{equation}\label{eq:gradient}
    \nabla \bar{d}_{*} = -\frac{1}{\lambda \bar{f}_{*}} \nabla \bar{f}_{*},
\end{equation}
where it can be seen that the gradient of $\bar{d}_{*}$ is in the opposite direction as the gradient of $\bar{f}_{*}$, subject to a scaling factor ${1}/{\lambda \bar{f}_{*}}$. 
However, because the Eikonal equation~\eqref{eikonal} requires that the magnitude of the gradient is normalised to 1, the scaling factor is unimportant, and we simply need to normalise the gradient of $\bar{f}_{*}$, and invert the sign. 

Based on the gradient~\eqref{eq:gradient}, the predictive variance can be calculated using a first-order approximation: 
\begin{equation}
    \mathbb{V}\left[d_{*}\right]=\frac{1}{\lambda \bar{f}_{*}} \mathbb{V}\left[f_{*}\right]\frac{1}{\lambda \bar{f}_{*}}^{\top}. 
\end{equation}
Despite its simplicity, a surprising benefit is that \eqref{3} approximately satisfies the Eikonal equation~\eqref{eikonal} for all query points.
In what follows, we present the derivation. 
}
\subsection{\edit{Derivation}}
A major challenge in exploiting the Eikonal equation~\eqref{eikonal} \edit{to recover the EDF} is that it is non-linear and hyperbolic. 
It is therefore difficult to solve directly, and special numerical approaches are often required, including discrete and combinatorial methods such as fast marching or label-correcting to propagate the distance field through a grid (\emph{e.g.}~\cite{Voxblox}~\cite{lau2010improved}). 

A work in the computer graphics literature~\cite{Crane} presents a smooth alternative based on Varadhan's distance formula~\cite{Varadhan}~\cite{Belyaev}, which approximates the EDF $d(\mathbf{x})$ using the heat kernel on the manifold $S$.
\edit{The physical intuition behind Varadhan's formula is as follows. 
Imagine that the target surface $S$ is hot and emanates heat. 
The conduction of heat can be viewed as particles taking a random walk starting from the boundary $\partial S$.
If we restrict the duration of the random walks to be short (\emph{i.e.} as time $t \rightarrow 0$), the paths taken will be close to the shortest possible one. 
}

\edit{Heat conduction on $S$ is modelled by the heat kernel, denoted by $v(\mathbf{x})$. 
The heat kernel is the solution of the \edit{heat} equation\footnote{Technically, this is a screened Poisson equation. Nonetheless, it exhibits similar behaviours to the conventional heat equation. We use the term heat equation in the interest of physical intuition.}:}
\begin{equation}
\begin{aligned}
    &   (1-t \Delta) v=(1/t- \Delta) v=0 \quad \text{ in } S\label{poissonE}, 
    \\
    &v=1 \quad \text { on } \partial S,
\end{aligned}
\end{equation}
where $\Delta = (\sum_{i} \frac{\partial^{2}}{\partial x_{i}^2}) $ is the Laplace operator and $t$ \edit{represents time}.
It is important to note that the \edit{heat} equation~\eqref{poissonE} is linear, unlike the Eikonal equation~\eqref{eikonal}. 

The celebrated result of Varadhan~(Theorem 2.3, \cite{Varadhan}) is that the heat kernel $v(\mathbf{x})$ and the EDF $d(\mathbf{x})$ are related as follows\footnote{Note that an analogous relationship also holds for the conventional heat equation (Theorem 2.2, \cite{Varadhan}), albeit with a more cumbersome counterpart of the log-transform.}:
\begin{equation}\label{limE}
    d(\mathbf{x})=\lim _{t \rightarrow 0}\{-\sqrt{t} \ln [v(\mathbf{x})]\}.
\end{equation}
where it can be seen that the approximation gets better as $t$ approaches zero \edit{(\emph{i.e.} after a short duration).}
Consequently, we define the approximation of $d(\mathbf{x})$ as $u(\mathbf{x})$:
\begin{equation}\label{logE}
    u(\mathbf{x})=-\sqrt{t} \ln v(\mathbf{x}).
\end{equation}
In this case, $t$ serves as a smoothing factor of $u(\mathbf{x})$. 
\edit{To check the effect of the approximation, we substitute $v(\mathbf{x}) = \exp(-u(\mathbf{x}) / \sqrt{t})$ into \eqref{poissonE}}:
\begin{equation}\label{eq:regularized_eikonal}
    v-t \Delta v=v\left[\left(1-|\nabla u|^{2}\right)+\sqrt{t} \Delta u\right]=0,
\end{equation}
which shows that we effectively solve the regularised form of the Eikonal equation.

While the theory described thus far only applies to the case of $\mathbf{x} \in S$ (due to the domain constraint in~\eqref{poissonE}), 
we can easily extend it to $\mathbf{x} \in \mathbb{R}^{D}$.
To do so, consider the EDF of the complement, $S^{c}$. 
The corresponding heat equation~\eqref{poissonE} is then defined in $S^{c}$.
Meanwhile, the boundary condition remains the same, because $\partial S^{c} = \partial S$. 
Combining the two, we have:
\begin{equation}
\begin{aligned}
    &v-t \Delta v=0 \quad \text{ in } \mathbb{R}^{D}\label{eq:heat_pde_euclidean}, 
    \\
    &v=1 \quad \text { on } \partial S,
\end{aligned}
\end{equation}
in place of~\eqref{poissonE}. 
Note that, in doing so, we have lost the information on whether $\mathbf{x}$ is inside or outside $S$ (\emph{i.e.} the sign is lost), because we have `stitched' the EDFs on $S$ and $S^{c}$. 
In our view, the loss of sign is a small price to pay compared to the convenience of using~\eqref{eq:heat_pde_euclidean} to approximate the (unsigned) distance field everywhere on $\mathbb{R}^{D}$. 


\edit{
Varadhan's formula~\eqref{limE} replaces the \emph{nonlinear} PDE constraint~\eqref{eikonal} on $d(\mathbf{x})$ with a \emph{linear} one~\eqref{eq:heat_pde_euclidean} on $v(\mathbf{x})$, which we can easily incorporate into GP regression.
Inspired by~\cite{sarkka_pde}, we do so by choosing a suitable covariance function using the Wiener-Khinchin theorem~\cite{GPbook}.
For brevity, we only discuss the 2D case and re-write \eqref{poissonE} as:
\begin{equation}\label{eq:heat_pde_2d}
    \frac{\partial^{2} v\left(x_{1}, x_{2}\right)}{\partial x_{1}^{2}}+\frac{\partial^{2} v\left(x_{1}, x_{2}\right)}{\partial x_{2}^{2}}-\lambda^{2} v\left(x_{1}, x_{2}\right)=w\left(x_{1}, x_{2}\right)
\end{equation}
where $w(x, y)$ is white noise and $\lambda = 1/\sqrt{t}$. The spectral density is the square of Fourier transform of~\eqref{eq:heat_pde_2d} as:
\begin{equation}\label{eq:spectral_density}
S\left(\omega_{1}, \omega_{2}\right)=\frac{1}{\left(\omega_{1}^{2}+\omega_{2}^{2}+\lambda^{2}\right)^{2}}\,.
\end{equation}
The Wiener-Khinchin theorem~\cite{GPbook} implies that the covariance function of GP satisfying~\eqref{eq:heat_pde_euclidean} is given by the inverse Fourier transform of spectral density~\eqref{eq:spectral_density}.
This is precisely the \edit{Whittle} covariance function\edit{~\cite{Whittle}, a special case of the Mat\'{e}rn family~\eqref{eq:mattern} with $\nu=1$}:
\begin{equation}
    k\left(\mathbf{x}, \mathbf{x}^{\prime}\right)=\frac{\left|\mathbf{x}-\mathbf{x}^{\prime}\right|}{2 \lambda} K_{\nu}\left(\lambda\left|\mathbf{x}-\mathbf{x}^{\prime}\right|\right)\,.
\end{equation}
}
\edit{Finally, the boundary conditions in~\eqref{eq:heat_pde_euclidean} translate to the choice of target measurements $y_{i} = 1$ at locations $\mathbf{x}_{i}$ on the surface. 
} 

\begin{figure}[b]
  \centering
  \resizebox{\linewidth}{!}{
  \subfigure[$\lambda = 5$]{\includegraphics[scale=0.23]{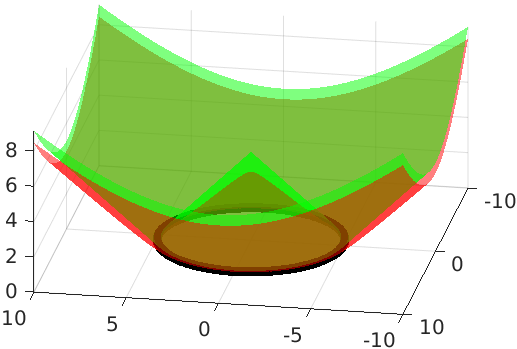}}
  \subfigure[$\lambda = 40$]{\includegraphics[scale=0.23]{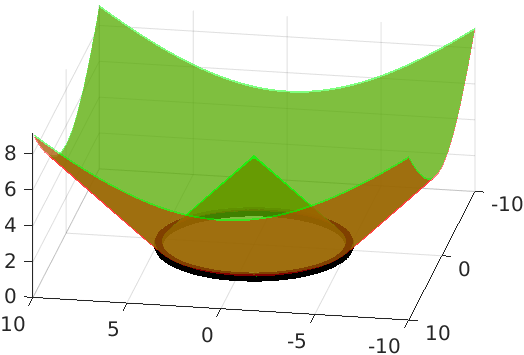}}
  }
  \caption{Whittle covariance approximation of the EDF in a 2D example. The measurements are the black points in a circle of $5$ m radius. The testing area is a square of $20 \times 20$ m. The difference of distance field between the prediction (shown as red color) and the ground truth (shown as green color) for two different $\lambda$ values. The larger the $\lambda$ the closer to the true distance.}
  \label{lambda}
\end{figure}

\subsection{Practical Considerations}
\edit{Note that when $\lambda$ is set to a large value, $t$ in \eqref{limE} gets closer to zero, thus producing a better distance approximation.} As shown in Fig.~\ref{lambda}, relatively large $\lambda$ values will produce a more accurate EDF. \edit{However, $t$ cannot be exactly zero as it is in the denominator in $v(\mathbf{x}) = \exp(-u(\mathbf{x}) / \sqrt{t})$. As a trade-off, we use $\lambda=40$ in our experiments.} In this paper, hyperparameter optimisation is not considered. Nevertheless, $\lambda$ is exactly the hyperparameter of the Whittle covariance, and \eqref{limE} provides the necessary intuition to choose a good $\lambda$ value for the EDF inference. 

\begin{figure}[t]
	\centering
	\resizebox{0.8\linewidth}{!}{
	\includegraphics[]{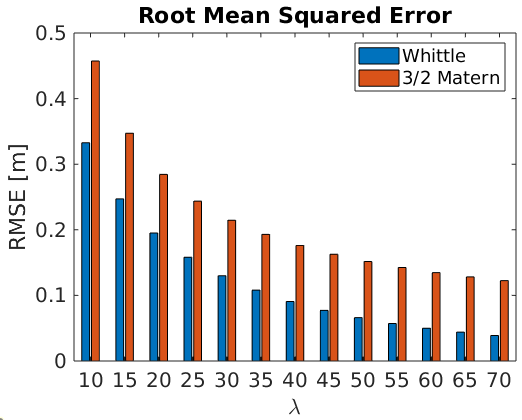}
	}
	\caption{RMSE varying $\lambda$ and covariances for the 2D example in Fig.~\ref{lambda}.}
\vspace{-2ex}
	\label{RMSE_kernel}
\end{figure}

\edit{As sample functions from Mat\'{e}rn family are $|\nu-1|$ times differentiable}, the Whittle covariance is $|\nu-1|=0$ times differentiable. This means it can not be used for GPIS gradient inference. Given that the Whittle covariance is a special case of the Mat\'{e}rn covariance family, and Mat\'{e}rn $\nu=3/2$ is once differentiable, the Mat\'{e}rn covariances are more suitable for distance estimation with gradients. 

We empirically observed that a modified version of Mat\'{e}rn $3/2$ with $l={\sqrt{2 \nu}}/{\lambda}$ is a good approximation of Whittle covariance with the advantage that it is once differentiable. As an example of this observation Fig.~\ref{RMSE_kernel} presents the root mean squared error in the case of different $\lambda$ values of both covariances for the example of Fig.~\ref{lambda} \edit{(see demo code}\edit{\footnote{\url{https://github.com/LanWu076/Log-GPIS-demo}}}). Given that the modified Mat\'{e}rn $3/2$ allows \edit{surface normal and distance gradient} prediction and Whittle does not, and the approximation is acceptable, we propose to use this Mat\'{e}rn version as a trade-off for extra information. 

\begin{figure*}[ht]
  \centering
  \resizebox{\linewidth}{!}{
  \subfigure[Ground Truth~\cite{Bhoram}]{\includegraphics[height=4cm]{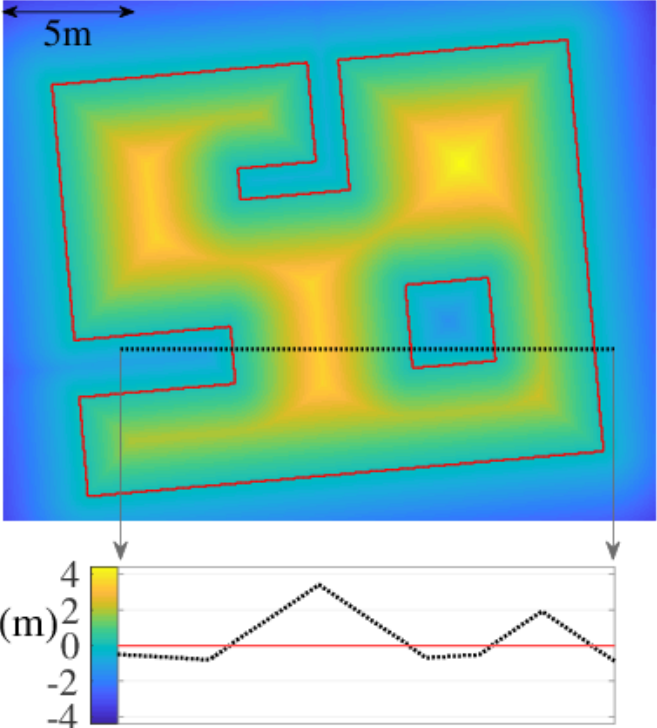}}
  \subfigure[TSDF~\cite{Bhoram}~\cite{kinectfusion}]{\includegraphics[height=4cm]{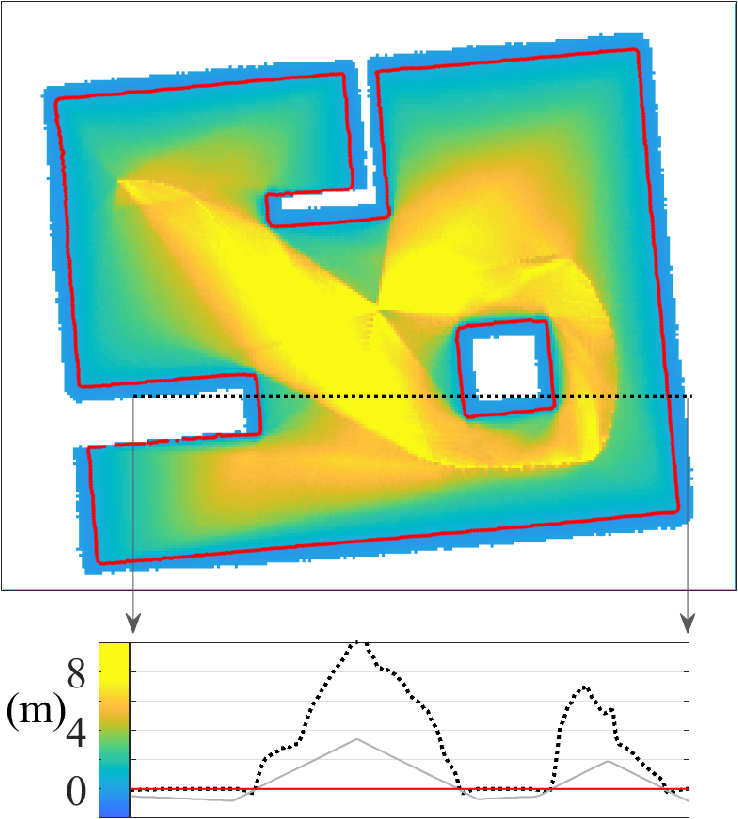}}
  \subfigure[GPIS-SDF~\cite{Bhoram}]{\includegraphics[height=4cm]{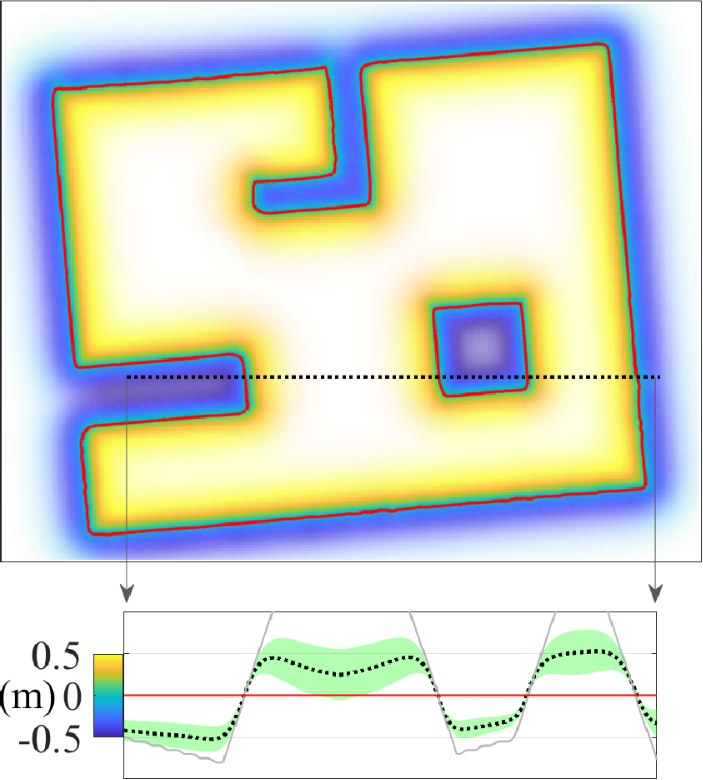}}
  \subfigure[Log-GPIS]{\includegraphics[height=4cm]{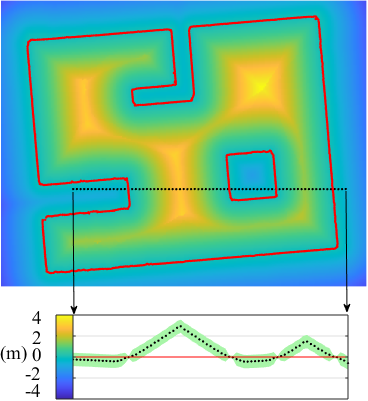}}
  }
  \caption{Distance field comparison of 2D map dataset. The 1D surface illustrates is shown in solid red line in all maps. The ground truth surface and ESDF map shown in a), then followed by the TSDF map in b) from the results presented in ~\cite{Bhoram}. In c) we present GPIS-SDF~\cite{Bhoram}, which is also based on a GPIS method. d) shows the results of our Log-GPIS method. All methods closely follow the surface, but only Log-GPIS infers the EDF accurately over the whole field. The bottom subplot show the distance value across the black line. Note that the colormap is different for the different maps visualisation purposes, except for the GT and ours. The light green region in c) and d) is variance.}
  \label{2D maps}
  \vspace{-2ex}
\end{figure*}

To summarise, Log-GPIS is capable of producing accurate EDF estimates by simply applying a log-transform to a GPIS with a covariance of the Mat\'{e}rn family. While the resulting EDF now loses the sign, it is possible to recover the sign using the surface normal and sensor's location, as we show in Fig.~\ref{2D maps}d). Moreover, there is no need to provide sample points inside and outside the surface. Our algorithm only requires the measurements around the surface, and then the EDF will be predicted through the queried points in the field.

\section{Evaluation}\label{sec:evaluation}
\subsection{Implementation}
We implemented the proposed Log-GPIS based on the open source implementation\footnote{\url{https://github.com/leebhoram/GPisMap}} of Lee \emph{et al.} in~\cite{Bhoram}. Our implementation maintains the online continuous mapping setup as the original code, with modifications that implement the Log-GPIS method described in the previous section to produce a faithful EDF of the scene.

It is well-known that GP-based methods are computationally expensive due to matrix inversion. The computational complexity scales as $\mathcal{O}\left(N^{3} D^{3}\right)$ with gradients, where $N$ is the number of training points and $D$ is the dimension of gradient. Moreover, the computation of predictive mean and variance costs $\mathcal{O}\left(N_{x} D\right)$ and $\mathcal{O}\left(N_{x}^{2} D^{2}\right)$ respectively per query point. Therefore, it is challenging to model large maps using most GP approaches. To address this practical issue, we adopt the \edit{incremental} approach of \cite{Bhoram}, where QuadTrees~\cite{finkel1974quad} and Octree~\cite{meagher1982geometric} data structures are used to maintain sub-GPIS of clusters, and online \edit{Bayesian update of surface points~\cite{Bhoram}} is used to deal with multiple overlapping observations. Furthermore, we modified the original code to render the final mesh using the marching cubes method~\cite{lorensen1987marching}. To increase the speed and accuracy of the extraction of marching cube vertices and faces, we generate the isosuface with Hermite data~\cite{fuhrmann2015accurate}. As discussed in Section~\ref{sec:approach}, we use Mat\'{e}rn $3/2$ covariance. The code is implemented in C++ and Matlab and runs on an Octa-core i7 CPU at 2.5GHz for all our experimental validations. 

In the following sections, we evaluate the performance of Log-GPIS in three scenarios. First, we show the comparison of the true distance value with ours and other approaches on a simulated 2D dataset from~\cite{Bhoram} with 2D laser observations. Subsequently, we quantitatively evaluate the accuracy of the predicted distance field in a 3D structure simulated in a Gazebo simulation environment with RGB-D image observations. Finally, in order to demonstrate the performance of our algorithm on a real dataset, we provide a quantitative and qualitative evaluation on a publicly available dataset that contains RGB-D images, ground truth sensor poses and structure ground truth. For all experiments, we represent 2D surfaces with continuous lines and 3D surfaces as meshes.\edit{\footnote{A video is available at \url{https://youtu.be/bSGx\_WNdQvo}\label{foot:video}}}
\subsection{2D Results in Simulated Dataset}
To ensure fair comparison between the results published in~\cite{Bhoram}, we first verify the accuracy of the Euclidean distance value by reproducing the same 2D simulation environment as in~\cite{Bhoram}. The dataset consists of \edit{$28$} Lidar scans from different robot poses with angular range of $-135^\circ $ to $ +135^\circ$\edit{, the sensing resolution of $1^\circ$, }and noise $\sigma=0.01m$. 

The results of TSDF~\cite{kinectfusion}, GPIS-SDF~\cite{Bhoram}, and our method Log-GPIS against the ground truth are shown in Fig.~\ref{2D maps}. The sub-figures at the bottom show the distance field along the solid black line. The ground truth in Fig.~\ref{2D maps}a) shows the linear behaviour of the distance to the surface. Given that this simulation shows the signed distance field, the surface is exactly at the zero crossing. Fig.~\ref{2D maps}b) shows the distance field estimated by the TSDF method of~\cite{kinectfusion} with constant weights. It turns out that the distance value of TSDF is overestimated within the limited viewpoints. It matches the true Euclidean distance only if the sensor ray is in the direction of surface normal. The benefit of TSDF is that it still produces an accurate continuous implicit surface. Fig.~\ref{2D maps}c) is the result of what the authors in~\cite{Bhoram} called GPIS-SDF, which is an online continuous and probabilistic method to recover the implicit surface and the distance field. Note that in this case although the distance value within a small region around the surface is accurate, if the testing points are far away from the measurements, the uncertainty increases and the distance values are not properly estimated. As we can see, the region without measurements nearby is covered by the white color with brighter being more uncertain. Fig.~\ref{2D maps}d) is the surface and distance field implemented by our method Log-GPIS. As we have mentioned previously, our approach does not contain signed distance information. However, we recover the sign of the Log-GPIS \edit{by comparing the predicted gradient of each testing point with sensor position. If the gradient is in the opposite direction of the sensor, we flip the sign of distance value of the testing point.} In this figure, it is evident that the EDF is linear and accurately predicted even far from the surface measurements, and the surface is accurately predicted. This shows that our Log-GPIS has the advantage of both mapping and distance accurate estimation. 

\begin{figure}[t]
	\centering
	\subfigure[Ground Truth]{\includegraphics[scale=0.27]{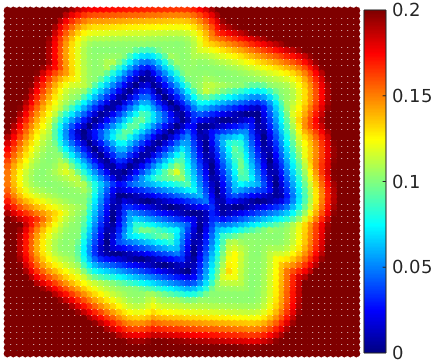}}
	\subfigure[Log-GPIS]{\includegraphics[scale=0.27]{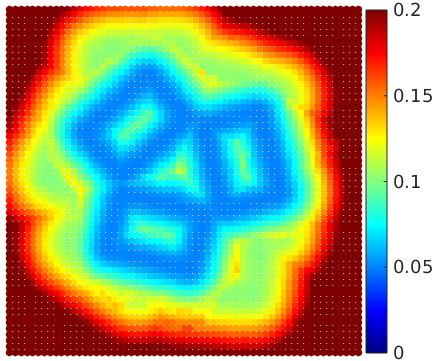}}
	\subfigure[Voxblox ESDF]{\includegraphics[scale=0.27]{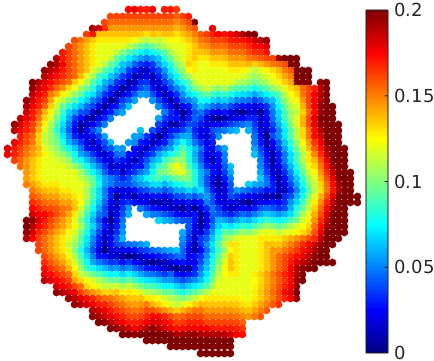}}
	\subfigure[GPIS-SDF]{\includegraphics[scale=0.27]{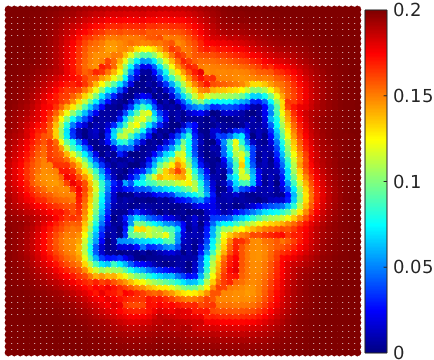}}
	\subfigure[Voxblox TSDF]{\includegraphics[scale=0.27]{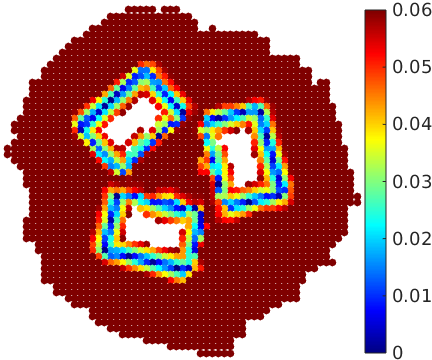}}
	\subfigure[RMSE of distance field]{\includegraphics[scale=0.235]{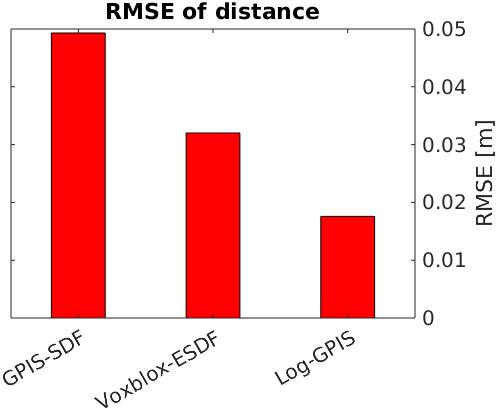}}
	\caption{Quantitative comparison of the distance field on a horizontal slice (size $1.2m\times1.2m$) of boxes dataset. The slice is 0.5 meter above the ground. All figures are from top view. It can be seen that Log-GPIS estimates the EDF with the lowest root mean square error.}
	\label{bricks ESDF}
\end{figure}

\begin{figure}[th]
	\centering
	\subfigure[Ground Truth]{\includegraphics[scale=0.27]{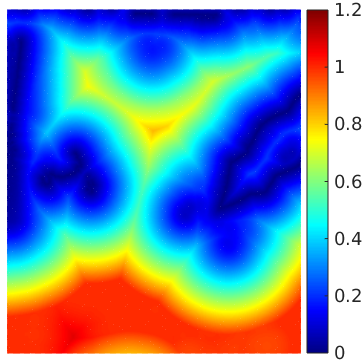}}
	\subfigure[Log-GPIS]{\includegraphics[scale=0.27]{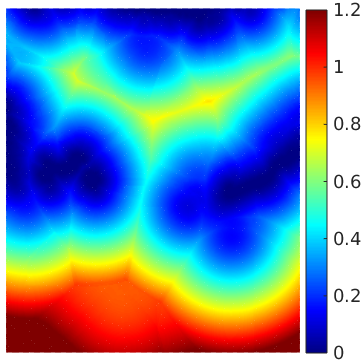}}
	\subfigure[Voxblox ESDF]{\includegraphics[scale=0.27]{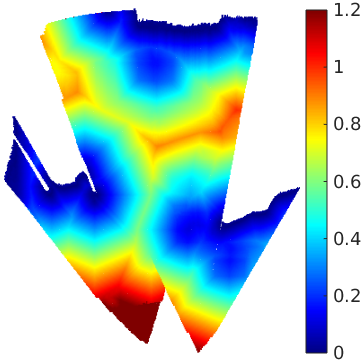}}
	\subfigure[GPIS-SDF]{\includegraphics[scale=0.27]{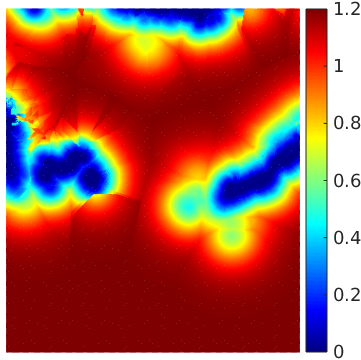}}
	\subfigure[Voxblox TSDF]{\includegraphics[scale=0.27]{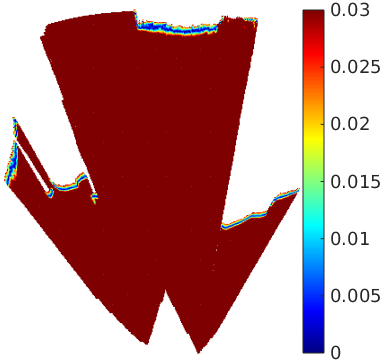}}
	\subfigure[RMSE of distance field]{\includegraphics[scale=0.235]{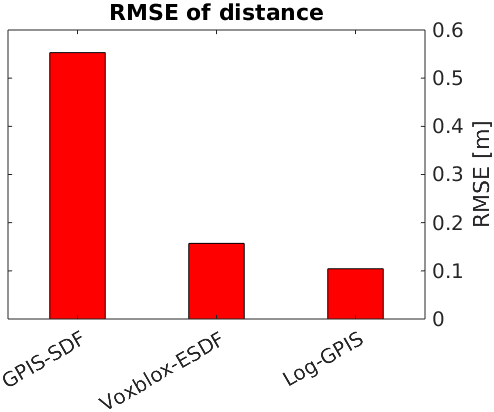}}
	\caption{Same quantitative evaluation as in Fig.~\ref{bricks ESDF}. The slice (size $3.75m\times3.2m$) is 1 meter above the ground. All figure are from top view.}
	\label{cow ESDF}
\end{figure}

\subsection{3D Results in Simulated and Real datasets}
In order to benchmark our method on 3D datasets, we present results from both simulation and real world data. 

The simulated dataset consists of a pile of boxes simulated in Gazebo. The simulation environment contains multiple boxes arrange in a three layer pile. The size of each box is $0.3\times0.2\times0.2m$. The raw measurements are 52 depth images captured by a robot with RGB-D camera moving around the boxes with accurate poses.

The real world experiment is based on the publicly available dataset \textsl{cow and lady dataset}~\footnote{https://projects.asl.ethz.ch/datasets/doku.php?id=iros2017}. The sensor measurements were obtained from a Kinect 2 RGB-D camera moving in 3D in a small room, with ground-truth poses measured by a Vicon motion capture system. The ground truth structure is a PLY Pointcloud, created from multiple scans with a Leica MS50 professional laser scanner. Since there is minor misalignment between each frame given the ground-truth poses, we use Iterative Closest Point algorithm~\cite{ICP} to correct the poses using the point clouds from consecutive frames.

We benchmark Log-GPIS with two state-of-the-art mapping frameworks; the recently proposed Voxblox~\cite{Voxblox} and the continuous mapping framework used in the previous section GPIS-SDF from~\cite{Bhoram}. These two frameworks are particularly relevant to our work as both produce a mesh reconstruction for visualisation purposes and generate concurrently the ESDF for navigation purposes. \edit{Given that~\cite{Voxblox} has validated the accuracy and efficiency of Voxblox against an occupancy representation, we directly compare Log-GPIS with Voxblox.} Due to our implementation is based on~\cite{Bhoram}, we simply use the same mapping parameters between GPIS-SDF and Log-GPIS. The parameters used for Voxblox were chosen to make the comparison as fair as possible. We chose a voxel size for simulation and real world datasets as 0.02 and 0.01 respectively, and the truncation distance as 3 voxels. We also chose the merged method as the TSDF Integrator to achieve the best TSDF and ESDF. Log-GPIS and GPIS-SDF testing points, and the Voxblox voxels are at the same resolution. 
\begin{figure}[t]
  \centering
  \resizebox{\linewidth}{!}{
  \subfigure[Voxblox mesh]{\includegraphics[height=3.8cm]{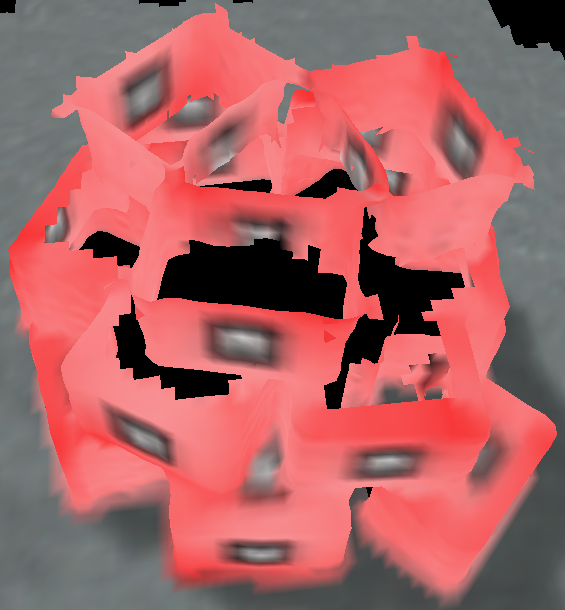}}
  \subfigure[Log-GPIS mesh]{\includegraphics[height=3.8cm]{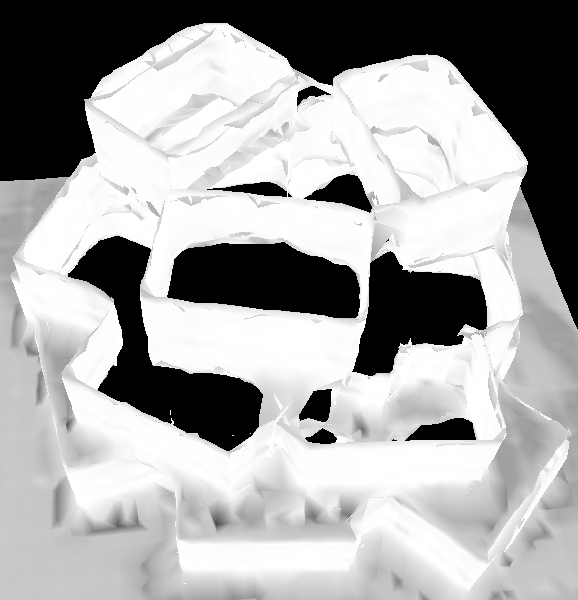}}
  }
  \resizebox{\linewidth}{!}{
  \subfigure[Log-GPIS error]{\includegraphics[height=3.8cm]{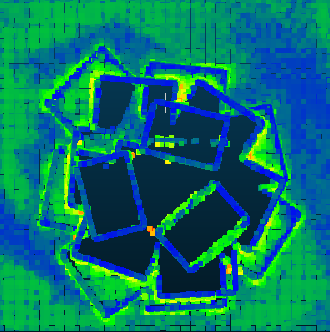}}
  \subfigure[Log-GPIS variance]{\includegraphics[height=3.8cm]{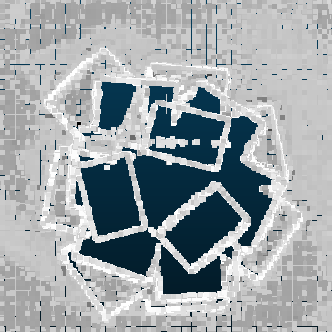}}
  }
  \resizebox{\linewidth}{!}{
  \subfigure[Voxblox error]{\includegraphics[height=3.8cm]{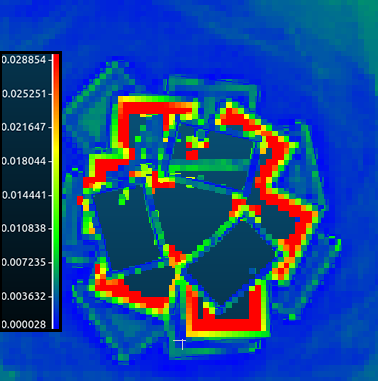}}
  \subfigure[Error of vertices]{\includegraphics[height=3.2cm]{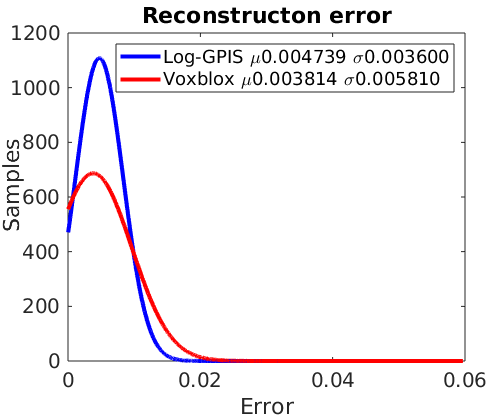}}
  }
  \caption{3D Reconstruction evaluation in boxes dataset. a) and b) show the meshes for Voxblox and Log-GPIS respectively with their associated reconstruction error in c) and e) and the error distribution in f).}
  \label{bricks mesh}
\vspace{-2ex}
\end{figure}
\begin{figure}[ht]
  \centering
  \subfigure[Voxblox Mesh]{\includegraphics[height=3.3cm]{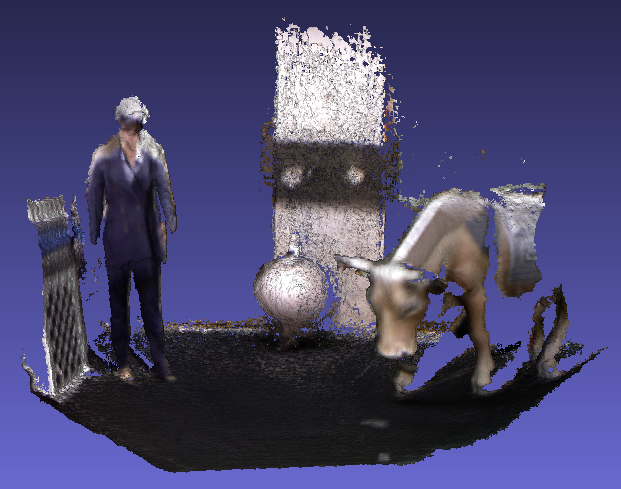}}
  \subfigure[Log-GPIS Mesh]{\includegraphics[height=3.3cm]{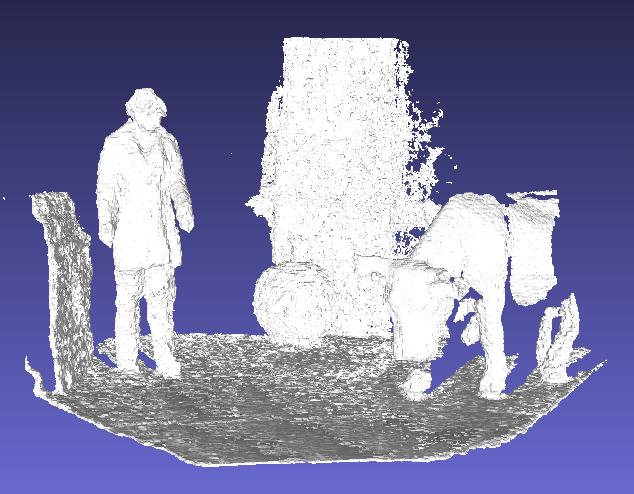}}
  \caption{3D reconstruction of the Lady and Cow dataset. a) Voxblox mesh with colour and b) Log-GPIS mesh coloured by variance.}
  \label{cow mesh}
\vspace{-2ex}
\end{figure}
\subsubsection{Accuracy of the Euclidean distance field}
To study the accuracy of distance field for the different methods, a cross-section of the simulated environment illustrating distance values is shown in Fig.~\ref{bricks ESDF} (see also Fig.~\ref{teaser}b)). Except for the colormap of the TSDF, the other sufigures are in the same colormap scale. To show clearly the edge of the brick in the boundary, we limit the colormap to a maximum of $0.2$ m for display purposes. In this case, we opted to compare the absolute distances, since the Log-GPIS explicity produces the unsigned distance. Fig.~\ref{bricks ESDF}a) presents the ground truth Euclidean distance computed from ground truth point cloud. The true surface is shown in black in the centre of the dark blue area. The ESDF and TSDF results from Voxblox are shown in Fig.~\ref{bricks ESDF}c) and e) respectively. Note that the field is only estimated by Voxblox in the areas within the field of view of the camera for the different poses. Voxblox constructs the TSDF first by grouped raycasting, and then propagating the ESDF based on the relatively accurate TSDF within truncation distance. By considering multiple observations of same region, the ESDF becomes more accurate. As we can see from the results, given the limited number of frames (52 depth images), the ESDF of Voxblox do not produce proper boundaries, especially the second layer of the boxes (shown in green colour) is not present. As for the TSDF, accurate distance values only within the vicinity of the hit surface. Fig.~\ref{bricks ESDF}d) presents the results of the distance field from GPIS-SDF. The distance follows a truncated distance field due to the lack of information far away from the surface. Fig.~\ref{bricks ESDF}a) presents the results of the proposed Log-GPIS EDF. In this case the distance follows very similar trend to the ground truth. Note that Log-GPIS instead of having zero-crossing on the surface, it is an inflection point (minimum value) at the surface. Our algorithm then finds the global minimum to recover the surface, which should be exactly in the middle of the light blue region. The accurate mesh estimation from Log-GPIS is shown in Fig.~\ref{teaser}b). As a quantitative measure we compute the root mean squared error for all the approaches. This is shown in Fig.~\ref{bricks ESDF}f). Note that for Voxblox, we only use the distance in the mapped space. The TSDF is not used in the RMSE evaluation since the distance is only valid inside the truncated region. While Voxblox-ESDF has a relatively good performance, our method produces the most accurate distance estimation.

We use the same benchmark in a more challenging scenario with the real data from the cow and lady dataset. Despite sensor noise and relatively inaccurate poses, the observations of the simulated dataset are similar in the real dataset as Fig.~\ref{cow ESDF} shows. Again the proposed Log-GPIS produces the most accurate distance field with the lowest RMSE amongst Voxblox ESDF and GPIS-SDF. 

\subsubsection{Surface reconstruction}
Here, we aim to evaluate the quality of the surface mesh using the same benchmarks as for the distance field. The ground-truth structure for the box dataset is shown in Fig.~\ref{teaser}a) and Fig.~\ref{bricks mesh} shows quantitative and qualitative results for Voxblox and Log-GPIS. Note that in this case GPIS-SDF and Log-GPIS produce nearly identical results for the reconstruction of the surface, thus we opted to omit the GPIS-SDF result.

Fig.~\ref{bricks mesh}a) and
b) present the 3D meshes of Voxblox and Log-GPIS respectively. In order to show the probabilistic information of Log-GPIS, we colored the surface with uncertainty, where darker areas relate to higher variance. Fig.~\ref{teaser}b) shows the colored version of the Log-GPIS mesh. Voxblox and Log-GPIS produce high-quality reconstructions. We note that the edges of Log-GPIS mesh are less noisy than Voxblox. Fig.~\ref{bricks mesh}c) and e) display the error evaluation of each mesh with respect the ground truth. The points in these two plots are coloured with same scale of colormap shown in Fig.~\ref{bricks mesh}e). Interestingly, the Log-GPIS uncertainty is well correlated with the error as shown Fig.~\ref{bricks mesh}d) and c) respectively. The overall error distribution displayed in Fig.~\ref{bricks mesh}f) shows that the mesh reconstruction performance for Voxblox and Log-GPIS is equivalent. The median computation time per frame of Voxblox including TSDF, ESDF and meshing is 1.54 s with the selected parameters. Our Log-GPIS consumes a median of 4.56 s including marching cubes for mesh generation\edit{, which allows online operations even with surface reconstruction.} Note that our method also allows partial evaluation at a faster computation time. Evaluating a horizontal line as a representative use-case for, \emph{e.g.} collision checking for navigation purposes, the median time is reduced to 3.21 s.

Finally, we present qualitative 3D reconstruction results for the real dataset in Fig.~\ref{cow mesh}. Voxblox reconstruction results are shown in Fig.~\ref{cow mesh}a) and the Log-GPIS colored by variance is shown in Fig.~\ref{cow mesh}b), where darker gray relates to high variance. We can see that both algorithms produce similar reconstruction results.

\section{Conclusion}\label{sec:conclusion}
We proposed a unified representation well-suited for robotic mapping and planning. The novel continuous and probabilistic Log-GPIS provides both, accurate EDF and implicit surface estimation with gradients. The derivation of this representation follows an elegant mathematical formulation exploiting the heat kernel method. The simple yet relevant modification of adding a logarithm transform to the well-known GPIS has the potential to enable wide adoption in practical mapping and planning applications. The results presented here validate the accurate performance of the Log-GPIS with respect to the state-of-the-art method in terms of EDF estimation. In future work, we consider to use our representation for collision check within a local planner. Further development aims at exploiting the provided continuous surface in an ICP-like framework to solve the simultaneous pose estimation and mapping problem.
\bibliographystyle{IEEEtran}
\bibliography{reference}
\end{document}